%% file: main.tex
\documentclass{article} % For LaTeX2e
\usepackage{iclr2025_conference,times}

\usepackage{hyperref}
\usepackage{url}

\title{CodeCipher: Learning to Obfuscate Source Code Against LLMs}
\author{%
  Yalan Lin\textsuperscript{\textnormal{1}} \hspace{.5em} Chengcheng Wan\textsuperscript{\textnormal{2}} \hspace{.5em} Yixiong Fang\textsuperscript{\textnormal{1}} \hspace{.5em}  Xiaodong Gu\textsuperscript{\textnormal{1}}\thanks{Xiaodong Gu is the corresponding author} \\[0.5ex]
  \textsuperscript{1}\,Shanghai Jiao Tong University ~
  \textsuperscript{2}\,East China Normal University\\[0.5ex]
  \scalebox{0.9}{\texttt{\{linyalan,xiaodong.gu\}@sjtu.edu.cn}}\\
  \scalebox{0.9}{\texttt{ccwan@sei.ecnu.edu.cn}}
}

\usepackage{booktabs}
\usepackage{tabularx}
\usepackage{amsmath,amssymb,amsfonts}
\usepackage{algorithmicx}
\usepackage[ruled]{algorithm2e}
\usepackage{algpseudocode}
\usepackage{graphicx}
\usepackage{subcaption}
\usepackage{longtable}
\usepackage{geometry}
\usepackage{adjustbox}
\usepackage{textcomp}
\usepackage{enumitem}
\usepackage{hyperref} % Load referencing package first
\usepackage{multirow}
\usepackage{xcolor}
\usepackage{colortbl}
\usepackage[skins]{tcolorbox}
\usepackage{xspace}

\newcommand{\ourmethod}{CodeCipher\xspace}

\usepackage{listings}
\usepackage{xcolor}
\usepackage{array}

\iclrfinalcopy % Uncomment for camera-ready version, but NOT for submission.
\begin{document}

\maketitle

\begin{abstract}
While code large language models have made significant strides in AI-assisted coding tasks, there are growing concerns about privacy challenges. The user code is transparent to the cloud LLM service provider, inducing risks of unauthorized training, reading, and execution of sensitive code. Such fear of data leaking prevents developers from submitting their code to LLMs. 
In this paper, we propose \ourmethod, a novel method that perturbs privacy from code while preserving the original response from LLMs. \ourmethod transforms the LLM's embedding matrix so that each row corresponds to a different word in the original matrix, forming a token-to-token confusion mapping for obfuscating source code. The new embedding matrix is optimized through minimizing the task-specific loss function. To tackle the challenge from the discrete and sparse nature of word vector spaces, \ourmethod adopts a discrete optimization strategy that aligns the updated vector to the nearest valid token in the vocabulary before each gradient update. 
We demonstrate the effectiveness of our approach on three AI-assisted coding tasks including code completion, summarization, and translation. Results show that our model successfully confuses the source code while preserving the original LLM's performance.\footnote{Code and data available at \url{https://anonymous.4open.science/r/CodeCipher_final-9D7E/}}
\end{abstract}

\section{Introduction}

The rise of code large language models (code LLMs), including CodeLlama~\citep{rozire2023codellama} and DeepSeek-Coder~\citep{deepseekcoder}, has ushered in a new phase for the research on code intelligence. These code language models have made significant strides in improving the quality of code generation and have achieved high pass rates across multiple benchmark datasets such as HumanEval~\citep{chen2021codex}, and MBPP~\citep{austin2021mbpp}. %For example, research shows that integrating Copilot leads to increased programmer productivity, with up to 8.91\% more weekly pull requests completed at Microsoft, highlighting its effectiveness in boosting coding efficiency. 
With the increasing reliance on these models for AI-assisted coding tasks, there is a growing need to leverage remote LLM cloud services, which enable broader scalability and enhance the capabilities of these models by offloading computations to powerful cloud infrastructure.

LLM cloud services allow users to perform AI coding tasks without the need for computational infrastructure. 
However, providing code LLMs as services raises significant privacy concerns, which limit their practical applications \citep{yao2024survey,yan2024protecting}. As users' code is transparent to these remote systems, there is a risk of exposing and misusing sensitive or proprietary information. Such fear of data leaking prevents many developers from submitting their code directly to LLMs. Therefore, privacy-preserving mechanisms in LLM deployment are required to ensure user trust and further expand code LLM usage.

%In previous work, two approaches have been explored to enhance the security of input code during the inference stage. %One approach employs cryptographic techniques, such as homomorphic encryption or multi-party computation. However, these methods typically necessitate modifications to the model architecture and entail high computational overhead, which in turn impacts the inference speed of the model. The second approach involves obfuscating the code. 

Code obfuscation techniques have been common solutions to defend code against unauthorized access, which disguises elements of a piece of code while maintaining its output. Existing techniques adopt rule-based perturbations, such as renaming variables with semantically meaningless symbols~\citep{jain2020contrastive} and injecting dead code~\citep{chakraborty2022natgen}, to obscure human comprehension of code semantics. However, since LLMs also rely on meaningful symbols and other human-readable code structures \citep{ding2024signaturecodesum,rodeghero2014improving}. Brute code obfuscation could significantly degrade the capability of LLMs and thus are inapplicable in practice.

In this work, we propose \ourmethod, a novel learning-based code obfuscation technique tailored for LLMs. \ourmethod safeguards code from unauthorized training, reading, compiling, and execution, without sacrificing LLM service quality. The core idea behind CodeCipher is to transform the LLM's embedding matrix so that each row corresponds to a different word in the original matrix. 
%the learned embedding matrix is discretized into tokens in the vocabulary using the original embedding matrix, yielding a word-to-word confusion mapping. 
This process creates a token-to-token confusion mapping, which the system uses to obfuscate tokens when encountering new code snippets.
%\wan{the method description is so hard to understand. Do you mean ``For each embedding layer of a given LLM, \ourmethod transforms its embedding matrix such that each row points to a different vocabulary word. \ourmethod then optimizes this new embedding matrix with a task-specific loss that xxxx, which forms the word-to-word confusion mapping that used for obfuscating tokens of the input code.''?}

The new embedding matrix is optimized by minimizing the task-specific loss. 
Due to the discrete and sparse nature of word embedding space, straightforward gradient descent can not pinpoint the embedding of a valid token. To tackle this problem, we adopt a discrete optimization strategy: at each iteration, the algorithm updates the perturbation vector, projects it onto the nearest valid token in the embedding space, and recalculates gradients based on the new projection. 
%\wan{The current writing is really hard to understand. I would prefer talking about the high-level idea and overall workflow of \ourmethod, and put the iterative design into the workflow.  Besides, can we move this challenge before introducing \ourmethod? so the logic flow is ``traditional method hurts model efficacy, straightforward optimization is ineffective. to tackle these problems, we proposes \ourmethod''}

The efficacy of our approach was rigorously assessed across three AI-assisted coding tasks: code completion, code summarization, and code translation. Results revealed that our method surpassed a range of conventional obfuscation methods in terms of both the level of confusion and the preservation of performance for downstream tasks. These findings affirm the practicability of our approach for enhancing privacy and security in LLMs. 

% \begin{figure}
%     \centering
%     \includegraphics[width=0.6\textwidth, trim=30 7 30 40 clip]{figures/intro.pdf}
%     \caption{Schematic illustration of our objective.}
%     \label{fig:intro}
% \end{figure}

% To summarize, our main contributions are twofold:
% \begin{itemize}[leftmargin=12pt]
%  \setlength\itemsep{0pt}
% %\item We conducted an evaluation of existing code obfuscation techniques and highlighted their inadequacy in adequately addressing the security concerns of users employing LLMs.
%     \item We proposes \ourmethod, the first code obfuscation technique tailored for LLMs, which protects code privacy without sacrificing LLM efficacy.
%     \item We evaluate \ourmethod with three representative AI-assisted coding tasks. The evaluation result shows that \ourmethod outperforms traditional code obfuscation methods.
% \end{itemize}

\section{Related Work}

\textbf{Privacy Protection for Large Language Models}
While LLMs have brought tremendous value to many fields, they have also sparked deep concerns among users regarding data privacy and security~\citep{nasr2023scalable}. 
Numerous studies have been conducted on model security in previous research \citep{fan2023fatellm,yan2024protecting,lin2024promptcrypt}.
% 模型端
On the model side, federated learning and differential privacy are commonly used to protect privacy leakage during model training. %Federated learning keeps data on local devices, with only model updates (gradients or parameters) being aggregated to a central server for model training \citep{fan2023fatellm}. 
%Differential privacy \cite{} adds random noise during the training process, ensuring that the model output does not change significantly due to the inclusion or exclusion of a single data point. 
Homomorphic encryption is widely used during model inference \citep{lin2024promptcrypt}. %For example, homomorphic encryption \citep{lin2024promptcrypt} allows computation on encrypted data directly, while multi-party computation \cite{} enables multiple parties to jointly compute a function without revealing their respective input data. Additionally, computation within an isolated Trusted Execution Environment (TEE) \cite{}. 
%用户端
On the user side, privacy protection focuses on marking or transforming user's data. Watermarking, the most common technology, embeds hidden markers in data to detect its use in model training \citep{sun2023codemark}.
Another method is data transformation, which makes data unintelligible or unusable even if leaked. Traditionally, this involves rule-based removal of sensitive information \citep{machanavajjhala2007diversity}. More recently, LLMs have been leveraged to obscure sensitive data \citep{song2024llm}. Users can construct privacy-preserving prompts, mixing real and fake inputs to prevent the server from identifying the true prompt \citep{utpala2023locally}. For example, TextObfuscator \citep{zhou2023textobfuscator} obfuscates input sentences at the embedding level, replacing tokens with semantically equivalent words to maintain model performance.

% While the aforementioned methods have contributed to the development of privacy protection for large models to some extent, a mature and widely-adopted solution has yet to be achieved. Each technology has its limitations, such as additional computational overhead, reduced model performance, and inevitably slower inference speeds \cite{}. Efficient and practical privacy protection mechanisms still require further exploration.

Our approach also focuses on data transformation but is specifically tailored for code scenarios. Unlike natural language, it is not just strings or prompts that need protection—the entire algorithm must be safeguarded. Rule-based removal of sensitive information is insufficient, and simply replacing variable names with synonyms, as done by TextObfuscator, leaves the code largely understandable, reducing the effectiveness of obfuscation.

% \subsection{Text Obfuscation for LLMs}

% \gu{discuss TextObfuscator (https://aclanthology.org/2023.findings-acl.337.pdf)}

% \lin{TextObfuscator \cite{zhou2023textobfuscator} also applies perturbations in the token embedding space to achieve privacy protection. However, it first fine-tunes the model using token prototypes to learn clustered representations, encouraging functionally similar words to converge around the same prototype during training. It then employs random perturbations to identify replacement words from those with similar representations. This approach can be computationally intensive when applied to LLMs, and collecting token properties is especially challenging in code-related tasks. Additionally, the paper aims to replace words with semantically or functionally similar vocabulary, which somewhat diminishes the effect of obfuscation.}

\textbf{Privacy Protection for Code LLMs}
Data leakage is also a concern in large language models for code (LLM4Code), which are often trained on open-source code repositories that may contain sensitive information. Studies~\citep{feng2022automated, huang2023not, al2023traces} have shown that specific prompts can extract private data like passwords and server keys from large models. %Some leaked information has been verified and poses real security risks~\cite{huang2023not}, indicating that the source of leaks is the training data, not the model generation itself.
Numerous solutions have been proposed to address this issue, such as adding watermarks to detect unauthorized use of training data \citep{sun2023codemark}, training models on desensitized data (though its impact on performance is still unknown) \citep{yang2024robustness}, and  distilling models to smaller, more secure local models \citep{shi2024greening}. However, ensuring the safe and private use of LLMs, especially for code generation, remains a critical challenge that requires further research.

\textbf{Code Obfuscation}
% 这里是否需要具体对混淆方法进行介绍
Code obfuscation aims to reduce the readability of source code through code transformation, reorganization, and modification \citep{behera2015different}. Commonly used code obfuscation techniques include identifiers renaming~\citep{jain2020contrastive}, control flow flattening \citep{laszlo2009obfuscating}, dead code injection \citep{chakraborty2022natgen}, self-modifying code \citep{giffin2005strengthening}, and property encryption \citep{pandey2012property}. 
The applications of code obfuscation are twofold. On the one hand, it makes the logic and structure of code harder to understand, complicating static and dynamic code analysis, and preventing attackers from assessing sensitive information or altering functionality. As a result, it is widely applied in software security to prevent reverse engineering, code theft, and vulnerability exploitation. On the other hand, code obfuscation can be viewed as introducing disturbance noise to the code data, and thus can be used as adversarial training samples to enhance the robustness of code language models \citep{jain2020contrastive,zhou2022adversarial}. 

While code obfuscation has been widely used for code security, its applicability to LLMs remains largely unexplored. Traditional rule-based methods achieve the goal of obfuscation, but they are not specifically tailored to LLMs. As a consequence, the code representations learned by LLMs can be altered, leading to significant performance degradation.

\section{Problem Statement}

%Given a code language model $P_\Phi(y|x)$ parametrized by $\Phi$, typically a GPT \cite{radford2019gpt2,mann2020gpt3} based on the Transformer architecture \cite{vaswani2017attention}. 
%Consider adapting this language model to downstream code intelligent tasks, such as code summarization, code generation, and code translation. 
% Each downstream task is represented by a training dataset of parallel pairs: $\mathcal{T}$= \{($x^{(i)}$, $y^{(i)}$)\} i=1,..,N, where both $x$ and $y$ are sequences of tokens. For example, in code summarization, $x$ is a code snippet and $y$ its corresponding natural language summary; for translation, $x$ is the code in a source language and $y$ its translation in the target language.\wan{I do not think these belongs to here. I created a shorter version in the problem statement. I think there should be a brief intro or high-level ideas of \ourmethod.}

%\wan{we may need a notation table}

%\subsection{Problem Statement}
Given a transformer-based code language model $P_\Phi(y|x)$ parameterized by $\Phi$ and a downstream task (e.g., code summarization) represented by a parallel dataset: $\mathcal{T}= \{(x^{(i)}, y^{(i)})\} $ $i=1,.., N$, where $x$ and $y$ are input and target sequences of tokens.
For each input source code \(x=\{w_1,...,w_n\}\), our goal is to transform it into obfuscated code \(x'=\{w'_1,...,w'_n\}\) which differs from \(x\) in lexicons while retaining task-specific performance (e.g., code completion) when processed by an LLM, formally:
%This endeavor is aimed at achieving both effective obfuscation and maintaining the functional integrity of the code.
\begin{equation}
    \min_{x'=g(x)}  ||s(P_\Phi(x'), y) - s(P_\Phi(x), y)|| 
    \label{eq:optimization}
\end{equation}
where $g:\mathcal{V}$$\rightarrow$$\mathcal{V}$ denotes a confusion function that maps any token in the vocabulary into an obfuscated one; $s:Y$$\times$$Y$$\rightarrow$$R$ denotes a task-specific scoring function such as Pass@K in the code completion task. %This obfuscation process can be implemented through human-defined rules (e.g., renaming identifiers into ordered symbols like $v_1$, $v_2$)\wan{no need to mention this. we are talking about an automated solution}. However, these mapping rules are primarily designed to obscure the understanding of code semantics by developers, which could hinder the LLMs in code comprehension as well. Furthermore, manually finding the optimal mapping without hurting the LLM performance is nontrivial\wan{The difficulty should be mentioned in introduction, not here}. 

\section{\ourmethod: Learning to Obfuscate Source Code}

\begin{figure}[htb]
    \centering
    \includegraphics[width=0.85\textwidth, trim=10 10 10 20 clip]{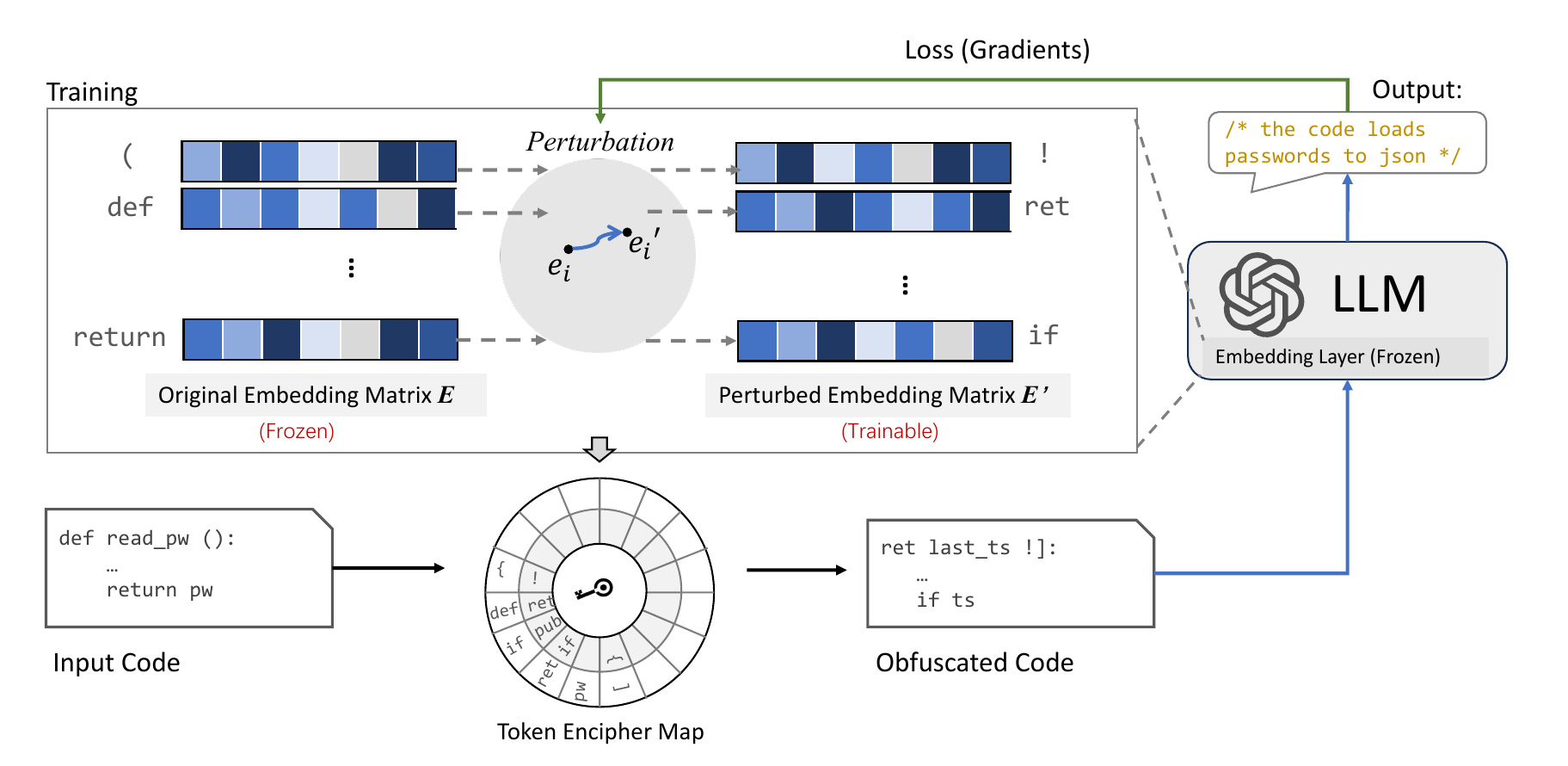}
    \caption{The overall framework of \ourmethod. It freezes the entire LLM and merely optimizes a perturbed embedding matrix $\mathbf{E}'$ for the embedding layer. %\wan{I recommand to highlight the updation in the figure, either through symbols, different fonts, or short phrases}
    }
    \label{fig:framework}
\end{figure}

In this paper, we propose a learning-based method to find the optimal confuse function \(g(\cdot)\) between code tokens. Unlike previous rule-based methods, we move our focus to the embeddings of code tokens. The core idea behind \ourmethod is to transform the LLM’s embedding matrix so that each row corresponds to a different word in the original matrix. The transformation establishes a token-to-token confusion mapping, which the system uses to obfuscate tokens when encountering new code snippets.
Our method consists of three stages as illustrated in Figure~\ref{fig:framework}: (i) transforms the LLM's embedding matrix to a permutation, establishing a token-to-token confusion mapping; (ii) optimizes the transformation of embedding by minimizing the task-specific loss function; addresses the intricacies involved in the discretization process, particularly the alignment of continuous word vectors to valid tokens; and (iii) obfuscates code using the trained model. A more comprehensive description of the algorithm can be found in Appendix A.

\subsection{Embedding Permutation}

For any code snippet \(x=(w_1,...,w_n)\), the embedding layer of an LLM converts the tokens into a sequence of continuous vectors:
  \begin{equation}
    \mathbf{e}_1,...,\mathbf{e}_T = \mathbf{E}(w_1), ..., \mathbf{E}(w_T)
  \end{equation}
where $\mathbf{E}\in\mathbf{R}^{|\mathcal{V}|\times d}$ represents the embedding matrix in the LLM's embedding layer, $|\mathcal{V}|$ denotes the vocabulary size, and $d$ represents the dimension of word embeddings.

\begin{figure}[htb]
    \centering
    \includegraphics[width=0.8\textwidth, trim=0 50 10 30 clip]{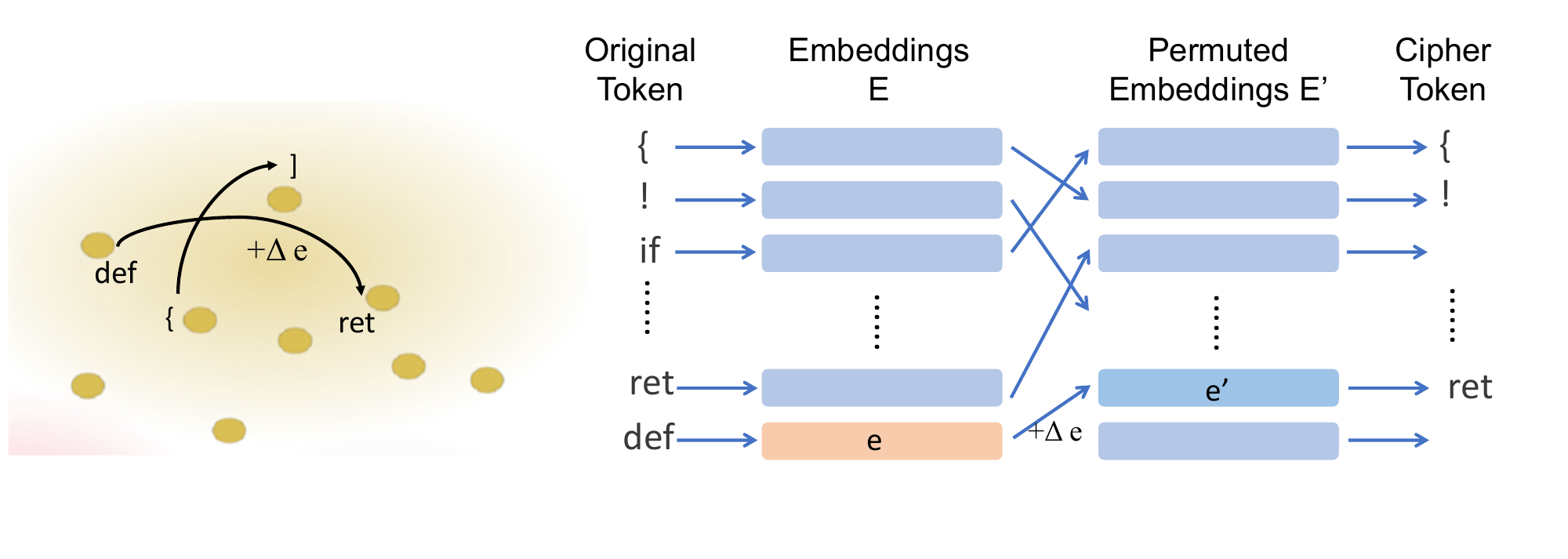}
    \caption{Illustration of Embedding Perturbation. The occurrence of token \texttt{def} can be reasonably substituted with \texttt{ret} through embedding transition, thereby obfuscating source code with these tokens.}
    \label{fig:emb_perturb}
\end{figure}

To perturb the code tokens, we create a learnable confusion mapping within their word embedding space, as illustrated in Figure~\ref{fig:emb_perturb}.
Specifically, given the original embedding matrix $\mathbf{E}$, we introduce a trainable embedding matrix $\mathbf{E}'$$\in$$\mathbf{R}^{|\mathcal{V}|\times d}$. The matrix $\mathbf{E}'$ is a permutation of $\mathbf{E}$, meaning that $\forall \mathbf{e}'\in\mathbf{E}'$, $\mathbf{e}'\in\mathbf{E}$. This ensures that every embedding in $\mathbf{E}'$ can be mapped back to its corresponding token in the vocabulary.
$\mathbf{E}$ and $\mathbf{E}'$ establishes a confusion mapping: each row $\mathbf{e}_i$ $\in$ $\mathbf{E}'$ represents the perturbed word vector corresponding to the original word vector in the same row of $\mathbf{E}$.

With the learnable confusion mapping, we design the obfuscation process as follows: for each token $w$ in the input code, we obtain its original embedding $\mathbf{e}=\mathbf{E}(w)$ using the embedding layer of the LLM. Then, we transform the embedding to $\mathbf{e}'=\mathbf{e}+\Delta \mathbf{e}$, where $\Delta \mathbf{e}$ denotes the perturbation of $\mathbf{e}$ in the corresponding row in $\mathbf{E}$.

The perturbed vector $\mathbf{e}'$ is then decoded to the nearest token in the vocabulary through a lookup function
$w'=\textsc{Dec}_\mathcal{V}(\mathbf{e}')$,
%$w'=\argmin_{v_i\in\mathcal{V}}\cos(\mathbf{e}',\mathbf{e}_i)$,
designated as the perturbed token. 
With this perturbation process, the entire sequence can be obfuscated as:
\begin{equation}
    x'= (w'_1, ..., w'_n) = \textsc{Dec}_{\mathcal{V}}(\mathbf{e}'_1, ...,\mathbf{e}'_n)
\end{equation}

\subsection{Training}

Our training objective is to find an optimal $\mathbf{E}'$ that minimizes the task-specific loss function $\mathcal{L}_\mathrm{task}$ when applied to the obfuscated code.
A straightforward idea is to optimize $\mathcal{L}_\mathrm{task}$ using gradient descent. 
For a code input, we update the embedding $\mathbf{e}$ for each token as follows:
\begin{equation}
\label{eq:gradient_update}
    \mathbf{e}' = \mathbf{e} - \eta \nabla_{\mathbf{e}} \mathcal{L}_{task}(x')
\end{equation}
where $\eta$ represents the learning rate, $\nabla_{\mathbf{e}}\mathcal{L}$ denotes the gradient of the loss function w.r.t. the current embedding $\mathbf{e}$. 

\begin{figure}
    \centering
    \includegraphics[width=0.7\linewidth]{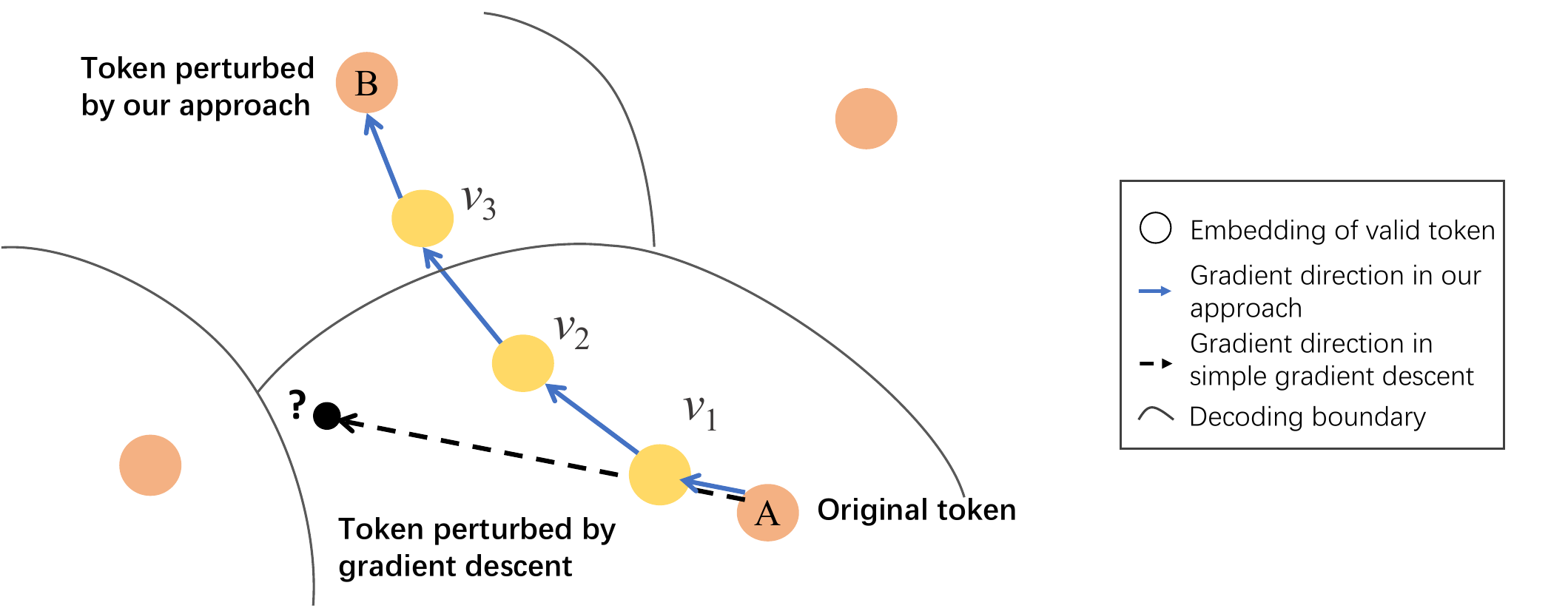}
    \caption{Illustration of discrete gradient search. Standard gradient descent perturbs token A along the gradient direction, often leading to suboptimal perturbations within the decoding boundary. Our discrete gradient search mitigates this issue by introducing mini-steps along valid tokens, ensuring more meaningful perturbations. %This exemplifies the difficulty in balancing the step size to achieve effective obfuscation without compromising the semantic integrity. %\wan{I also strongly recommend to keep figure 2 and 3 in the same style, and put in the same line with minipage}
    }
    \label{fig:iter}
\end{figure}

However, direct gradient descent encounters the intrinsic limitation in the word embedding space. 
%Although gradient descent can straightforwardly find the optimal embedding, it is nontrivial to identify the optimal obfuscation token. 
Code tokens reside on a sparse, discrete manifold in the embedding space. As illustrated in Figure~\ref{fig:iter}, a single gradient computation is inadequate for pinpointing the exact position of a token in the embedding space, thereby hindering precise vector manipulation. 
%\wan{Do we really need to emphasis such difficulty? Based on the current description, it seems to be a problem shared by any language model with embedding layers}

%Inspired from discrete optimization algorithms \citep{yuan2021bridge,wen2024hard}, 
%To tackle this challenge, 
Inspired from \cite{yuan2021bridge}, we develop a discrete gradient search algorithm. During each gradient update, the algorithm aligns the updated vector with semantically meaningful tokens, ensuring transitions occur between valid tokens in the original embedding space. More specifically, the algorithm performs gradient update in Equation \ref{eq:gradient_update} along a finer-grained trajectory~$v_1,...,v_t$. At each mini step, the embedding vector is incrementally perturbed and then projected onto the nearest valid token vector in the vocabulary. The gradient is then recalculated based on this new token, guiding further refinements in the following iterations. Formally, 
\begin{equation}
    v_{t+1} = \Pi_{\mathcal{V}}(v_t-\eta\nabla_{v_t} \mathcal{L}_\text{task}(v_t))
\end{equation}
% \begin{align}
%     v_t' & = \Pi_{\mathcal{V}}(v_t) \\
%     v_{t+1} & = {v}_t - \eta \nabla_{{v}_t'} L_\text{task}({v}_t')
% \end{align}
where ${v}_t$ denotes the token vector at the $t^{th}$ step; $\Pi_{\mathcal{V}}$ denotes the projection operation which aligns the updated vector onto the nearest word's vector within the vocabulary space $\mathcal{V}$; $\eta$ represents the learning rate; $\nabla_{{v}_t}\mathcal{L}({v}_t)$ denotes the gradient of the loss function $\mathcal{L}$ with respect to ${v}_t$.

This iterative process continues for a certain number of steps, at which point we select the vector with the lowest task loss and perform the final projection, designating it as the perturbed token. 

To prevent over-obfuscation, we limit the training process to a maximum of $N$ iterations. The training early stops if the perplexity deviation of a code sample reaches a predefined threshold $\delta$. We define $\delta$ as a linear function that increases with each step $i$, namely, $\delta = \alpha \cdot i + \beta $, where $\beta$ indicates the initial threshold and $\alpha$ controls the rate at which the threshold increases throughout the training process. % we employ a greedy update approach: once a token has been updated, it is marked as ``frozen," allowing the focus to shift toward optimizing the remaining unfrozen tokens in the vocabulary. After several iterations, the approach effectively obfuscates a significant portion of the tokens in the vocabulary. 

\subsection{Code Obfuscation}

Once the new embedding matrix is trained, it establishes a confusion function that maps each token in the vocabulary to its obfuscated counterpart. For any new input code snippet, \ourmethod converts the tokens by referencing this mapping table, producing an obfuscated version of the code. The obfuscated code can then be sent to a cloud LLM with the same specifications as the one used for training.

% \begin{figure}
%     \centering
%     \includegraphics[width=0.5\textwidth, trim=0 70 0 80 clip]{figures/changed.pdf}
%     \caption{Transformation of the entire vocabulary}
%     \label{fig:changed}
% \end{figure}

\section{Experiments}
\label{sec:downstream_task}
% 这里不确定三个任务是分开来写还是合并

\subsection{Common Setup}
As a proof of concept, we build and evaluate \ourmethod with CodeLlama-7B \citep{rozire2023codellama}, a representative white-box code LLM. We will show in Section~\ref{ss:exp:trans} that the obfuscated code generated by CodeLlama-7B can be transferred to a wide range of model specifications including different sizes and black-box LLMs. More details about the hyperparameter setting can be found in Table~\ref{tab:hyperparam} in the Appendix.

We train and evaluate our obfuscation model on three AI-assisted coding tasks.

\textbf{Code Completion.} 
A task that completes the subsequent code given users' partial code. 
We train our model on the MultiPL-E dataset (the Python subset from the Leetcode section) \citep{cassano2023multipl}, which comprises 446 code snippets. We test the model on the HumanEval benchmark \citep{chen2021codex} which contains 164 Python programming problems. For each problem, we obfuscate the first half of the code, provide it to the LLM, and ask it to complete the remaining half. As the obfuscated half code might prevent the final output from being compiled, we guide the model to re-generate the entire code from head using the prompt in Appendix~\ref{appendix:prompt}.\\ 
\textbf{Code Summarization.}
A task that provides a succinct natural language summary for a given code fragment~\citep{sridhara2010towards}. %Code summarization is widely used to validate the model's ability in code understanding~\citep{,,}. 
We train and test our obfuscation model on the CodeSearchNet benchmark~\citep{husain2019codesearchnet} using the prompt in Appendix \ref{appendix:prompt}. \\
\textbf{Code Translation.}
A task that translates source code written in one language to another \citep{pan2023lostintranslation}. %Code translation requires LLMs to acquire proficient knowledge in both languages: a deep understanding of the intent of the original while generating code in the target language.
We train our model on the Java-to-Python subset of XLCoST~\citep{zhu2022xlcost} %\lin{some codes only have java version and do not have corresponding python version, so we filter out code snippets written in only Java or Python}.
and test it on the Java-to-Python subset of HumanEval-X \citep{zheng2023codegeex}.
We use google-java-format\footnote{https://github.com/google/google-java-format} to ensure consistent line breaks and indentation across the code, aligning it with the formatting of the test dataset. This process resulted in a total of 2,374 data pairs. In the prompt in Appendix \ref{appendix:prompt}, we explicitly instructed the model to avoid class-based code structures.

\textbf{Baselines.}
We compare \ourmethod with both rule-based and LLM-based code obfuscation approaches.
1)~\textit{Random perturbation}, which randomly replaces a portion of tokens in the code with non-semantic symbols. 
2)~\textit{Identifier renaming}, which replaces variable and function names with non-semantic symbols~\citep{chakraborty2022natgen}. %Identifier renaming has been widely used for code obfuscation~\cite{chakraborty2022natgen} and is an effective way to protect the sensitive information contained in the variables from being leaked. We employ an identifier obfuscation tool named Spoon \cite{pawlak:hal-01169705} to replace variables with random strings (e.g. ved3kd3r2e). The original tool is written in Java. We wrote a script to port it into Python. 
It is implemented from Spoon~\citep{pawlak:hal-01169705}.
%\textbf{Operators\& Operands Swap.} Human beings tend to follow certain habits and norms when writing code, for example, variables are often placed on the left side of symbols and values are placed on the right side when writing expressions, we exchange the left and right parts of these expressions, and also change the method of symbols to reduce the readability of the code. We used scripts in NatGen\citep{} to do this.
3)~\textit{Dead branch injection}, which inserts code blocks that never execute (e.g., \texttt{while(false)}). %The method is effective in confusing and misleading the readers in comprehending the code, without changing the actual execution results. We used the scripts in NatGen \citep{chakraborty2022natgen} for the implementation.
It is implemented from NatGen~\citep{chakraborty2022natgen}.
4)~\textit{Remove symbols}, which removes formatting and syntactic symbols such as whitespace and parentheses from the code. %Since the obfuscated code will be re-tokenized by the LLM tokenizer, we evaluate whether LLMs can maintain the ability to comprehend the obfuscated code. 
%\wan{Is it a rule-based solution or LLM-based? please make it clear}
5)~\textit{Prompting LLMs}, %Since LLMs have the ability to understand human intentions and execute commands, we attempt to 
which prompts GPT-4o directly to obfuscate code using the prompt in Appendix~\ref{appendix:prompt}.
 %In contrast, for the various versions of CodeLlama, upon receiving the above prompt, they only output the original source code or modified code with altered indentation, and are unable to perform any deeper level of obfuscation.
6)~\textit{Obfuscation + Inform}, which obfuscates identifiers in the source code and explicitly informs the LLM about this obfuscation in the prompt (e.g., ``Generate a docstring for the code where the identifiers are obfuscated."). Previous work has shown that LLMs can better handle obfuscated code when they are made aware of the obfuscation upfront \citep{2021Towards}. %By doing so, the LLM can focus more on the structural aspects of the code, potentially leading to improved performance despite the obfuscation.

\subsection{Performance Preservation}
We measure the efficacy of \ourmethod based on two competitive objectives: task performance (i.e., Pass@k~\citep{chen2021codex}, k=1,10,100) and obfuscation degree (including \textit{perplexity} measured by the LLM and \textit{edit distance} to the original code). 
We start by presenting the efficacy in performance preservation, in other words, how the code obfuscated by \ourmethod maintains the performance of downstream tasks. 
Table \ref{tab:result:completion} shows the results of various methods in the code completion task\footnote{We show results of other tasks in Appendix}. To ensure a fair comparison, we maintained a consistent level of obfuscation across all approaches and compared their accuracy on downstream tasks. The results indicate that \ourmethod effectively confuses source code without substantially compromising the original model's performance. For instance, ``identifier renaming'' and ``random'' perturbation drop Pass@1 from 50.60\% to 42.68\% and 40.24\%, respectively, whereas our approach only reduces it to 47.56\%. 
The results demonstrate that our method achieves a superior balance between code obfuscation and model performance compared to the baselines, making it more practical in real-world applications.

\begin{table*}
\centering
\small
\caption{Results of Performance Preservation on the Code Completion Task.}%which measures the proportion of generated code that passes the test cases among $k$ generated results.}
\begin{tabular}{lccccccc}
    \toprule
    \multirow{2}{*}{\bf Method} & \multicolumn{3}{c}{\bf Task-specific Performance} & & \multicolumn{2}{c}{\bf Obfucation Degree} \\
    \cline{2-4}\cline{6-7}
     & \bf Pass@1(\%) &\bf Pass@10(\%) & \bf Pass@100(\%) &  &\bf PPL & \bf Edit distance(\%) \\
    \midrule
    Origin & 50.60 & 68.68 & 79.27 & & 3.56 & 0 \\
    Random perturb & 40.24 & 61.34 & 75.00 & & 5.95 & 3.86 \\
    \hline
    \bf Rule-based Obfuscation &   &   &   &   &  &\\
    Identifier renaming & 42.68 & 61.15 & 75.00 & & 5.64 & 8.74 \\
    Dead branch injection & 39.63 & 60.23 & 71.95 & & 4.71  & 9.21 \\
    Remove symbols & 37.20  & 60.93 & 75.60 & &\bf 6.09  & 3.38 \\
    \hline
    \bf LLM-based Obfuscation &   &   &   &    & &   \\
    Encipher with LLM prompting & 39.02  & 60.33  & 72.56 & & 4.57  & 8.94 \\
    Obfuscation + Inform & 41.46  & 57.67  & 70.12  & & 5.64  & 8.74 \\
    \ourmethod (ours) &\bf 47.56 &\bf 65.20 &\bf 77.15 &  &\bf 6.09 &\bf 9.41 \\
    \bottomrule
\end{tabular}
\label{tab:result:completion}
\end{table*}

%\subsection{Results under varying confusion levels}

\begin{figure}[htb]
  \centering
  \subcaptionbox{Pass@1 under different edit distances for code completion}{
    \includegraphics[width=0.3\textwidth, trim=50 20 0 0 clip]{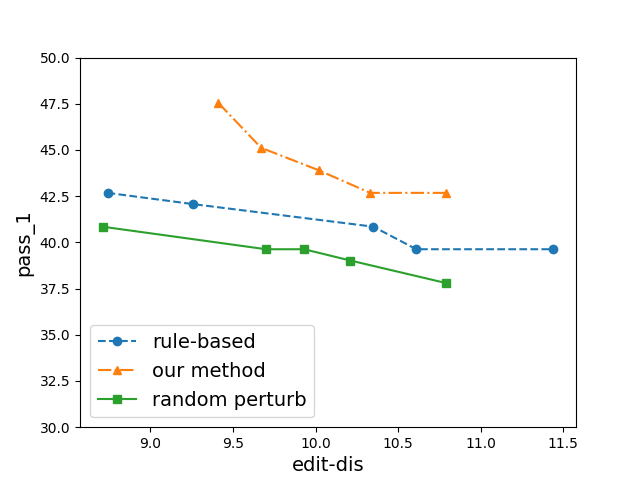}
  }
  \subcaptionbox{BLUE scores under different edit distances for code summarization}{
    \includegraphics[width=0.3\textwidth, trim=50 20 0 0 clip]{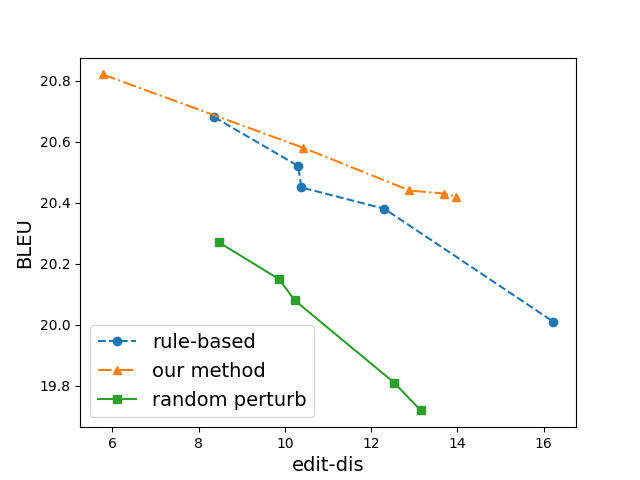}
  }
  \subcaptionbox{Pass@1 under different edit distances for code translation}{
    \includegraphics[width=0.3\textwidth, trim=50 20 0 0 clip]{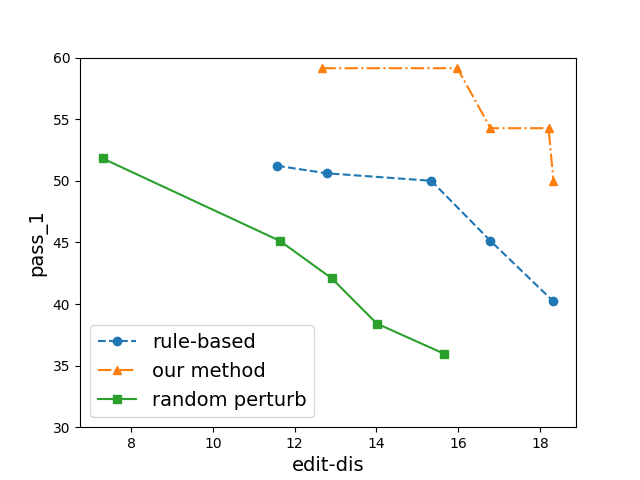}
  } \\
    %  %\hfill
  % \subcaptionbox{Pass@1 under different PPLs on code completion}{
  %   \includegraphics[width=0.3\textwidth, trim=50 20 0 0 clip]{figures/pass_1_gen_ppl_1.png}
  % }
  %   %\hfill
  %\subcaptionbox{BLUE scores under different PPLs on code summarization}{
  %  \includegraphics[width=0.3\textwidth, trim=50 20 0 0 clip]{figures/bleu_summary_ppl.png}
  %}
  %  %\hfill
  % \subcaptionbox{Pass@1 under different PPLs on code translation}{
  %   \includegraphics[width=0.3\textwidth, trim=50 20 0 0 clip]{figures/pass_1_translate_ppl.png}
  %}
  \caption{Task performance under various confusion levels}
  \label{fig:results:curve}
\end{figure}

%In the previous tables, we show results with a certain level of obfuscation across all approaches. 
To comprehensively assess \ourmethod under different obfuscation levels. We vary the confusion degree of various methods and compare their task performance. We compare \ourmethod with random perturbation and identifier renaming, two strong baselines in the previous results. %We control the obfuscation degree of our approach by varying the number of training steps, while for baseline models, 
We control the obfuscation degree by varying the proportion of perturbed tokens in the entire code. The results are presented in Figure \ref{fig:results:curve}. Across all tasks and metrics, our method exhibits consistent strength over baselines under different obfuscation levels. The results suggest that our method strikes a more effective balance between code obfuscation and maintaining performance on downstream tasks.

\subsection{Privacy Protection}
%\gu{Compare with baselines the accuracy of deobfuscation, the compilation rate, and the runnability rate, }

Having established that \ourmethod effectively preserves the task performance, we now assess its efficacy in privacy protection, the primary objective of code obfuscation. We aim to determine whether the obfuscation can obscure the information, %can obscure the semantic cues, 
and prevent the code from compiling, execution, and model training.

\textbf{Setup}:
We conduct experiments on the CodeSearchNet dataset which was collected from real-world projects in GitHub and may involve more private information. The raw dataset consists of 2 million samples. We randomly sampled 200 code samples and obfuscate them using various methods. 
We adopt three metrics: \textit{1) Compilation Rate}, the proportion of compilable code among all obfuscated code,
% A code snippet is considered compilable if it can be parsed into ASTs by treesitter. 
\textit{2) Deobfuscation Rate}, which measures the proportion of obfuscated code that can be deobfuscated by LLMs. 
We compute this proportion using the recall score, i.e., proportion of tokens in the original code that also appear in the deobfuscated code.
\textit{3) Deobfuscation Distance}, refers to the edit distance that the deobfuacated code deviate from the original code. For ease of comparison, we also compute the edit distances for the initially obfuscated code. 
We deobfuscate the obfuscated code by prompting CodeLlama-7b-Instruct. The prompt used for deobfuscation is provided in Appendix~\ref{appendix:prompt}.

\textbf{Results}:
As can be seen from Table \ref{tab:result:privacy}, \ourmethod disrupts the compilability of all obfuscated code, thereby preventing its reuse. 
Particularly, the obfuscated code is challenging for LLMs to restore, with only 34\% of tokens successfully restored. The same phenomenon can be seen from the deobfuscation distance. 
For all methods, the edit distances of the deobfuscated code are greater than those of the initially obfuscated code, probably because the deobfuscation process introduces additional modifications to the code. Notably, the code obfuscated by \ourmethod exhibits the largest edit distance from the original code compared to other methods. The results suggest that our approach successfully obfuscates code and enhances privacy protection.

\input{tables/results_privacy_csn_display}

\subsection{Ablation Study}

In this section, we conduct an ablation study on the discrete gradient search, a key step in our algorithm. We replace the discrete gradient search with traditional gradient descent, namely, a single gradient descent followed by decoding. We also vary the learning rates of gradient descent, a critical factor in our approach. As shown in Table \ref{tab:result:ablation_study}, relying solely on a single-step gradient descent remarkably affects the code completion results. Traditional gradient descent results in suboptimal performance under all learning rates, indicating the efficacy of discrete gradient search in our method. %Our findings indicate that symbols within the code are frequently the first elements altered. Perturbations in this part lead to an increase in the generation of structurally incorrect code, thereby affecting the final accuracy. 
\begin{table}
\centering
\small
\caption{Ablation study of the discrete gradient search on the code completion task}
\begin{tabular}{lcccccc}
    \toprule
    \bf Model &\bf Pass@1(\%) &\bf Pass@10(\%) & \bf Pass@100(\%) & \bf PPL & \bf Edit distance(\%) \\
    \midrule
    \ourmethod & 47.56 & 65.20 & 77.15 & 6.09 & 9.41 \\
    -\space w/o discrete gradient search (lr=5$e$-3) & 39.63  & 57.25  & 67.68 & 5.07  & 2.70   \\
    -\space w/o discrete gradient search (lr=1$e$-2) & 39.02  & 58.70  & 73.17 & 5.62   &  3.88 \\
    -\space w/o discrete gradient search (lr=2$e$-2) & 43.90  & 63.52  & 77.04 &  5.66  & 4.42  \\
    \bottomrule
\end{tabular}
\label{tab:result:ablation_study}
\end{table}

\subsection{Case Study}
\begin{figure} [htbp]
\centering
\subcaptionbox{A case for code completion. We obfuscate the first half of the original code and ask the LLM to regenerate the entire code given the obfuscated part.} {
    \includegraphics[width=0.9\textwidth, trim=50 0 50 0 clip]{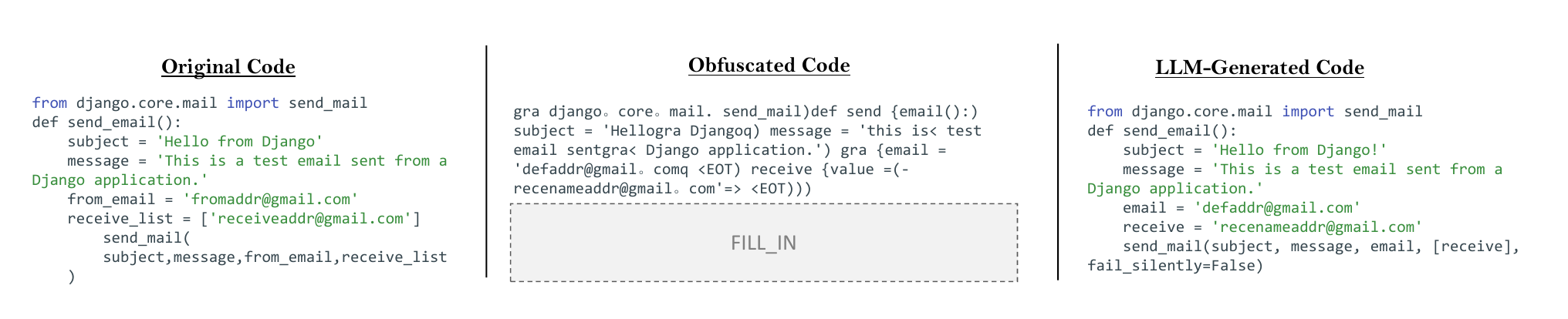}
}
\subcaptionbox{A case for code summarization. We obfuscate the entire code and compare the LLM-generated summaries for the original and the obfuscated code.}{
    \includegraphics[width=0.9\textwidth, trim=65 10 90 0 clip]{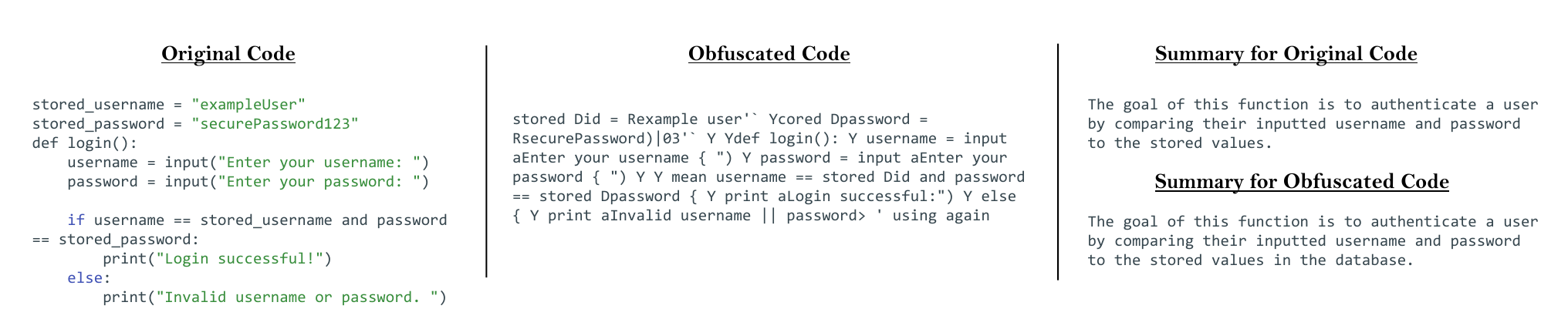}
}
\caption{Examples of code obfuscation by \ourmethod}
\label{fig:cases}
\end{figure}

Figure~\ref{fig:cases} presents two practical examples by \ourmethod.
We observe that \ourmethod obfuscates sensitive information such as email address and passwords in the original code. The privacy contained therein is protected. Nonetheless, the model remains capable of understanding the code and generating correct but insensitive responses.
In summary, the results show that our approach effectively obfuscates the sensitive information to LLMs, while preserving its functional correctness. 

\subsection{Transferability}
\label{ss:exp:trans}
Our approach is developed on white-box LLMs where the embedding layers and task gradients are accessible. In practice, there is always a need to apply the model in a new LLM, particularly black-box LLMs which provide only an API interface to users. We are unable to fine-tune these models based on their parameters. In this section, we aim to evaluate whether the obfuscated code generated by our method can still achieve good performance on other models.

\textbf{Setup:}
We used the obfuscated code generated by CodeLlama-7B with a perturbed embedding layer to evaluate the performance on other models. We conducted the experiments on the code completion task, which is one of the most common and general tasks. The representative models chosen for evaluation include StarCoder \citep{li2023starcoder}, DeepSeek-Coder \citep{deepseekcoder}, and GPT-3.5-turbo \citep{gpt35turbo}, which have shown strong performance on AI coding tasks. We employed greedy decoding during inference, and the HumanEval dataset was used for testing.

\begin{figure}[htbp]
    \centering
    \includegraphics[width=0.55\textwidth, trim=20 10 30 30 clip]{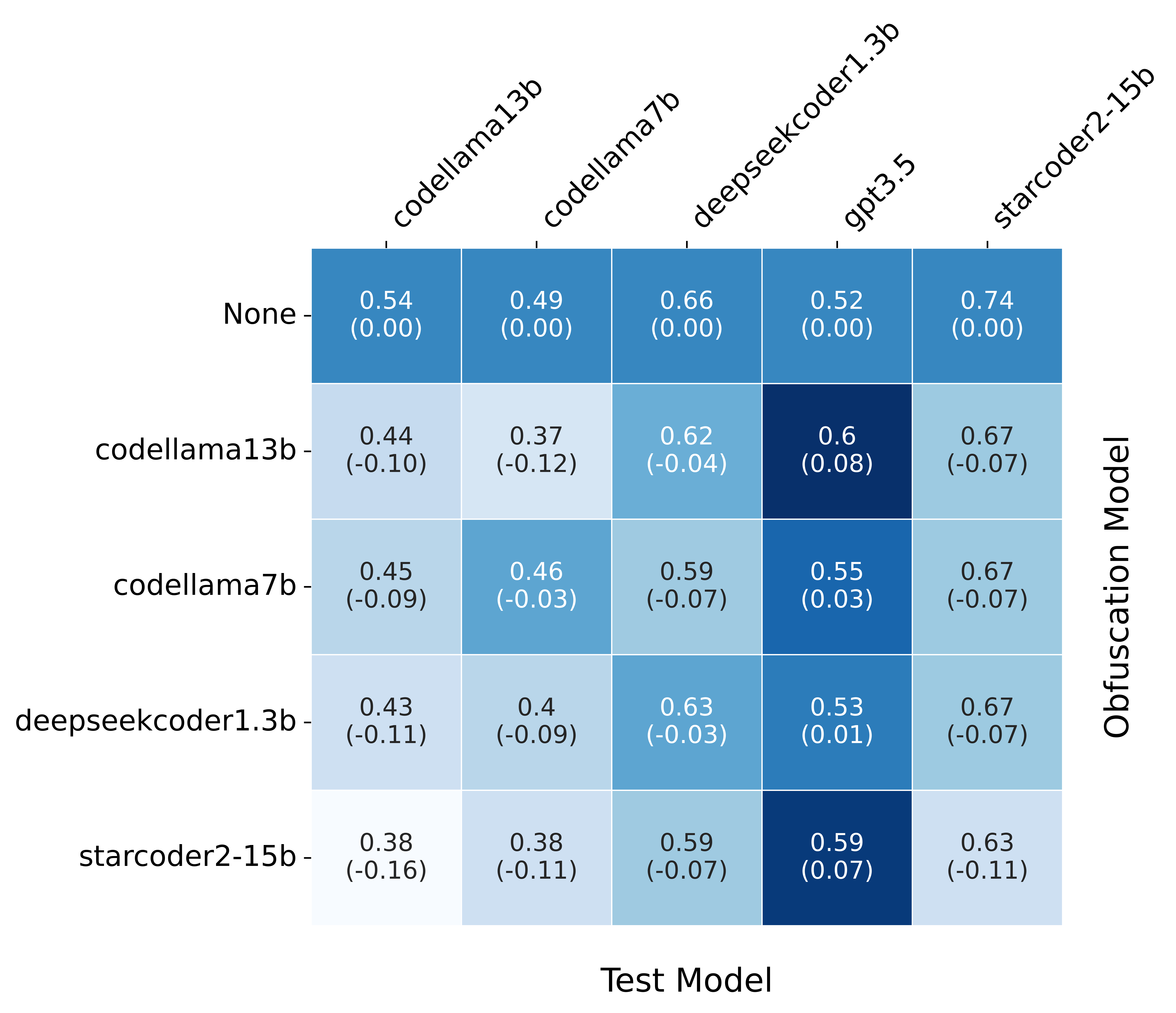}
    \caption{Results of the transferability test. The vertical axis represents different obfuscation models while the horizontal axis represents different test models. In each cell, the first line indicates the Pass@1 value achieved by the test model under the obfuscation model, and the value in parentheses shows the difference compared to the Pass@1 result without any obfuscation.}
    \label{fig:transfer}
\end{figure}

\textbf{Results:}
As shown in Figure~\ref{fig:transfer}, our method obtains relatively stable Pass@1 scores in each model through different obfuscating models, and in most cases, the deviation from original scores is within 10\%. In particular, in the black-box LLM GPT-3.5-turbo, the obfuscated codes outperform its original results, showing that our method remains high efficacy on other models. We hypothesize that our method learns common token mappings that are portable across LLMs. Overall, the results suggest that our method is not tied to any specific version of the LLM; instead, it is a general approach that works independently of the LLM version used. In practical applications, we can develop our method on a local proxy model such as the open-sourced CodeLlama-7B and apply the learned token confusion mapping to other cloud LLM servers. 

\section{Conclusion}
In this paper, we propose \ourmethod, a novel code obfuscation method tailored for LLMs. \ourmethod learns a transformation on the word embedding space, establishing a confusion mapping between code tokens that can be used for code obfuscation. The embedding transformation is optimized by minimizing the task-specific loss. A discrete gradient search algorithm is designed to tackle the sparsity challenge in the embedding space. Experiments on three AI-assisted coding tasks demonstrate that our method achieves a superior balance between code obfuscation and model performance compared to traditional rule-based methods.

%\subsubsection*{Acknowledgments}
%Use unnumbered third level headings for the acknowledgments. All acknowledgments, including those to funding agencies, go at the end of the paper.

\bibliography{references}
\bibliographystyle{iclr2025_conference}

\appendix

\section{The Algorithm}

The detailed algorithm for our approach is presented in Algorithm~\ref{algo}

\begin{algorithm}[h]
\small
\SetAlgoLined
\KwIn{Input-output pairs for a specific task: $\langle X, Y\rangle$ \\ 
$\quad \quad \quad $ The original embedding layer $\mathbf{E}$ of the LLM\\ 
$\quad \quad \quad $ Maximum training iterations $N$, PPL thresholds $\alpha$, $\beta$, Maximum steps for each iteration $T$\\
$\quad \quad \quad $ Learning rate $\eta$
}
\KwOut{The new embedding layer $\mathbf{E}'$ of the LLM}
$\mathbf{E}' = \mathbf{E}$ \\
\For{$i =1\, to\, N$}{
    $(x_i$, $y_i)\sim\langle X, Y\rangle$  
                    \Comment{Randomly choose a training sample $(x_i$, $y_i)$}\\
    $\mathbf{e} = \mathbf{E}'(x_i)$ \quad $x_i' = \text{Dec}(\mathbf{e})$ \Comment{Obfuscate the code using the previous confusion matrix}\\
    \tcp{Early stop if the code is sufficiently confusing}
    $\delta_{PPL} = \mathrm{PPL}(x_i') - \mathrm{PPL}(x_i)$ \\
    \If{$\delta_{PPL} \leq \alpha * i + \beta$}{
        $\mathbf{e}_{best}=\mathbf{e}$, $\mathcal{L}_\mathrm{best}=\mathcal{L}_\mathrm{task}(x'_i,y_i)$ \\
        \For{$t = 1$ \KwTo $T$}{
            $\mathbf{e}' = \Pi_{\mathcal{V}}(\mathbf{e})$ \quad
            $x_i'$ = Dec ($\mathbf{e}'$) \Comment{Decode embedding to vocabulary space}  \\
            
            $\mathbf{e} = \mathbf{e} - \eta \nabla_{\mathbf{e}'}\mathcal{L}_{\mathrm{task}}(x'_i, y_{i})$ \Comment{Gradient update w.r.t. the projected embedding} \\
            
            \If{$\mathcal{L}_\mathrm{task}(x'_i, y_i)<\mathcal{L}_\mathrm{best}$}{
                $\mathbf{e}_{best} = \mathbf{e}$, $\mathcal{L}_\mathrm{best} =  \mathcal{L}_\mathrm{task}(x'_i, y_i)$ \Comment{Record the best $\mathbf{e}$ that leads to the minimum loss}
            }
        }
        $\mathbf{e}' = \Pi_{\mathcal{V}}(\mathbf{e}_{best})$ \Comment{Final projection}  \\
        $\mathbf{E}'[x_i] = \mathbf{e}'$ \Comment{Replace the corresponding token embedding in $\mathbf{E}'$ using $\mathbf{e}'$}\\
    }
}
\Return $\mathbf{E}'$

\caption{\textbf{\ourmethod Algorithm}}
\label{algo}
\end{algorithm}

\section{Prompt Templates}
\label{appendix:prompt}

\begin{tcolorbox}[title=Prompt for Code Completion]

Please complete this code from head: 

\{code\}

Please only output the code. Please complete this code from head:

\end{tcolorbox}

\begin{tcolorbox}[title=Prompt for Code Summarization]

Generate a docstring for the code in 10 words. 

\{code\} 

Please fill this sentence: `The goal of this function is to' in 10 words

\end{tcolorbox}

\begin{tcolorbox}[title=Prompt for Code Translation]

Please translate this Java code to Python: 

\{code\} 

Please use a functional programming style

\end{tcolorbox}

\begin{tcolorbox}[title=Prompt GPT-4o to perform code obfuscation]

 Obfuscate the following Python function to make it difficult for humans to understand while keeping the normalized edit distance WITHIN 0.10. 
 Do not add or remove line breaks or blank spaces in the code. 
 Please only output the code. 
 
 \{Original-Code\}

\end{tcolorbox}

\begin{tcolorbox}[title=Prompt for Code Deobfuscation]

I have an obfuscated code (remove symbol) for you to restore, and your output should only be the code, and the code should be in ``` and ```. Please ONLY output the code, DO NOT output other things. The Obfuscated code is:

\{code\}

Please restore the original code:

\end{tcolorbox}

\section{Hyperparameter Setting}

The hyperparameters for the three tasks are provided in Table~\ref{tab:hyperparam}.

\begin{table} [h]
\centering
\small
\caption{Hyperparameter Setting.}
\begin{tabular}{lccc}
    \toprule
    Hyperparameter & Code Completion & Code Summarization & Code Translation \\
    \midrule
    Learning rate ($\eta$) & 0.002  &  0.002 &  0.002 \\
    Max sample number (N) & 32  &  32 &  32\\
    $\alpha$ &  1.5 & 1.5  & 2.0 \\
    $\beta$ &  $\frac{1}{90}$ &  $\frac{1}{20}$ &  $\frac{1}{90}$ \\
    Maximum iterations per session (T) & 10  &  10 & 10  \\
    \bottomrule
\end{tabular}
\label{tab:hyperparam}
\end{table}

\section{Results on other tasks}
Tables~\ref{tab:result:sum} and \ref{tab:result:trans} show the results on the code summarization and translation tasks, respectively. We observe the same trend as in the code completion task: \ourmethod increases the perplexity and edit distance to the original code while achieving higher performance in the downstream tasks. 
In code summarization, our approach attains a PPL that is 2.73 times higher than the original while keeping the reduction in BLEU scores within 2\%.
In code translation, the PPL of our obfuscated code is three times higher than the original, yet the Pass@1 decreases by only 4\%, and the Pass@100 score remains unchanged. 
These results confirm the efficacy of our approach in code obfuscation.

\begin{table} [h]
\centering
\small
\caption{Results on Code Summarization. We employ BLEU-4~\citep{papineni2002bleu}, ROUGE-L \citep{lin2004rouge}, and METEOR \citep{banerjee2005meteor} as the performance measures.}
\begin{tabular}{lcccccc}
    \toprule
        \multirow{2}{*}{\bf Method} & \multicolumn{3}{c}{\bf Task-specific Performance} & & \multicolumn{2}{c}{\bf Obfucation Degree} \\
    \cline{2-4}\cline{6-7}
     & \bf BLEU &\bf ROUGE-L & \bf METEOR &  &\bf PPL & \bf Edit distance(\%) \\
    \midrule
    Origin & 21.00 & 22.50 & 34.95&  &  6.52 &  0 \\
    Random perturb & 20.08 & 19.75 & 30.50 & &  28.05 & 10.23 \\
    \hline 
    \bf Rule-based Obfuscation &   &   &  & &   &   \\
    Identifier renaming & 20.47 & 20.50  &  32.55 & & 17.79 & 10.27 \\
    Dead branch injection & 20.57  & 19.00  & 32.71 & & 8.14 & 10.25 \\
    Remove symbols & 20.00  & 21.25  & 32.86 & & 18.02  & 10.34\\
    \hline
    \bf LLM based Obfuscation &  &   & &  & & \\
    Encipher with LLM prompting & 20.03  & \textbf{21.75} & 20.99 &  & 9.27 & 10.39  \\
    Obfuscation + Inform & 19.80  & 20.50 & 30.95 & & 17.79 & 10.27  \\
    \ourmethod (ours) & \textbf{20.58} & \textbf{21.75} & \textbf{33.34} &  & \textbf{34.03} & \textbf{10.42} \\
    \bottomrule
\end{tabular}
\label{tab:result:sum}
\end{table}

\begin{table} [h]
\centering
\small
\caption{Results on Code Translation. We employ the widely used Pass@k~\citep{chen2021codex} (k=1,10,100) as the performance measures.}
\begin{tabular}{lccccccc}
    \toprule
    \multirow{2}{*}{\bf Method} & \multicolumn{3}{c}{\bf Task-specific Performance} & & \multicolumn{2}{c}{\bf Obfucation Degree} \\
    \cline{2-4}\cline{6-7}
     & \bf Pass@1(\%) &\bf Pass@10(\%) & \bf Pass@100(\%) &  &\bf PPL & \bf Edit distance(\%) \\
    \midrule
    Origin & 61.59 & 85.90 & 92.68 & & 2.05 & 0 \\
    Random perturb & 51.83 & 74.65 & 87.80 & & 6.10 & 7.30 \\
    \hline
     \bf Rule-based Obfuscation &  &  &  &  & &  \\
    Identifier renaming & 50.61 & 82.00 & 90.85 & & 3.37 & 11.63 \\
    Dead branch injection & 49.39  & 75.29  & 86.59  & &  6.52 & 10.05 \\
    Remove symbols &  47.50 & 76.88  & 85.76  & & 6.25  & 3.79 \\
    \hline
    \bf LLM-based Obfuscation &   &   &   &   & & \\
    Encipher with LLM prompting & 57.31  & 82.55  & 92.07 &  &  3.09 &  10.36 \\
    Obfuscation + inform & 44.51  &  76.91 & 91.46 & & 3.37  & 11.63 \\
    \ourmethod (ours) & \bf 59.15 & \bf 83.97 & \bf 92.68 &  & \bf 6.53 & \bf 12.68 \\
    \bottomrule
\end{tabular}
\label{tab:result:trans}
\end{table}

\end{document}

%% file: tables/results_privacy_csn_display.tex
\begin{table}
\centering
\small
\caption{Performance on Privacy Protection on the CodeSearchNet Benchmark}
\begin{tabular}{lcccc}
    \toprule
    \multirow{2}{*}{\bf Method} & \bf Compilation & \bf Deobfuscation  & \multicolumn{2}{c}{\bf Deobfuscation Distance (\%)} \\
    \cline{4-5}
     & \bf Rate (\%) &\bf Rate (\%) & \bf Before deobfus. &\bf After deobfus. \\
    \midrule
    Random & 6 & 35   & 10.23 & 39.89    \\
    Identifier renaming & 76 & 38 &  10.27 & 39.09   \\
    Dead branch injection & 96 & 35 & 10.25  & 37.77   \\
    Remove symbols & 0 & 36 & 10.34 & 38.91  \\
    LLM prompting & 84 & 32  &  10.39 & 39.44   \\
    Obfuscation+Inform & 76 & 38 &  10.27 & 39.09  \\
    \ourmethod (ours) & 0 & 34 & 10.42  & 43.18 \\
    \bottomrule
\end{tabular}
\label{tab:result:privacy}
\end{table}